\documentclass[letterpaper, 10 pt, conference]{ieeeconf}  
\usepackage{graphicx}
\usepackage{empheq}
\usepackage{subfig}
\usepackage{multirow}
\usepackage{booktabs,siunitx}
\usepackage{comment}
\usepackage{rotating}
\usepackage{fancyhdr}
\usepackage[sorting=none]{biblatex}
\usepackage{amsmath}
\DeclareMathOperator*{\argmax}{argmax}

\newcommand{\niton}{\not\owns}

\IEEEoverridecommandlockouts                              

\title{\LARGE \bf
Pose-free object classification from surface contact features in sequences of Robotic grasps
}

\author{Teresa Alves$^{1}$, Alexandre Bernardino$^{2}$ and Plinio Moreno$^{2}$
\thanks{*This work was not supported by any organization}
\thanks{$^{1}$ MSc Student Finalist at Instituto Superior Técnico,
        Lisbon, Portugal
        {\tt\small teresa.alves@tecnico.ulisboa.pt}}%
\thanks{$^{2}$Researcher at the Institute for Systems and Robotics in Instituto Superior Técnico, Lisbon, Portugal}%
}

\begin{document}

\setcounter{page}{1}

\maketitle
\thispagestyle{plain}

\begin{abstract}

 In this work, we propose two cost efficient methods for object identification, using a multi-fingered robotic hand equipped with proprioceptive sensing. Both methods are trained on known objects and rely on a limited set of features, obtained during a few grasps on an object. Contrary to most methods in the literature, our methods do not rely on the knowledge of the relative pose between object and hand, which greatly expands the domain of application. However, if that knowledge is available, we propose an additional active exploration step that reduces the overall number of grasps required for a good recognition of the object. One of the methods depends on the contact positions and normals and the other depends on the contact positions alone. We test the proposed methods in the \textit{GraspIt!} simulator and show that haptic-based object classification is possible in pose-free conditions. We evaluate the parameters that produce the most accurate results and require the least number of grasps for classification.

\end{abstract}

\section{Introduction}

Recent improvements in tactile sensing and their implementation on humanoid robot hands is allowing new robots to emerge with new features and mimic different human activities and actions \cite{evolution}. In the past years, robots have become very good at grasping in highly controlled environments, like a factory assembly line \cite{dexterityMasters}, in tasks like picking up an object at an exact position and placing it in another exact position. The real world, however, is much more flexible and the need arises to automatize certain tasks, such as pick, identify and move some specific object. The idea for this project is to simulate the trivial human task of blindly identifying an object within a given set, through the object's haptic properties \cite{haptic-1993}.

We propose two different approaches for object identification by using a robot hand with some form of proprioception and/or tactile sensing. One of the methods relies on the ability to measure the 3D positions and surface normals at the finger contact points, in a hand centered reference frame. We denote this method \textbf{PN} (Point and Normal) based. The second method only uses the positions of the contacts, and not the contact surface normals, in order to account for the existence of more limited sensors. We denote this method \textbf{P} (Point) based. Experiments are made with a robotic simulator \cite{graspit} with the robotic Barrett Hand.

In the scenario where the relative pose between the robot hand and the object is available, we investigate an active learning strategy for exploration around the object to find the relative pose leading to the grasp with the highest information gain \cite{active-touch},\cite{activeLearning}.  We compare active exploration results with passive/random exploration (the base algorithm where the next grasp is chosen randomly) under the same conditions. The passive exploration approach consists in randomly exploring the object.

The validation of the proposed methods is validated in the \textit{GraspIt!} simulator \cite{graspit}. The robustness of the method is assessed by adding random noise to the acquired sensor values and checking the corresponding performance. The method keeps, at each grasp, a likelihood score for each object, so that a decision can be made at any time. With more grasps, the certainty on an object increases and we can stop the algorithm based on a threshold on the level of certainty, which leads to a decision on the object identity. We evaluate the method in terms of efficiency - how many grasps are required to reach a certain level of certainty - and accuracy - what is the fraction of correct decisions made by the system.


\section{Related Work}

There are already several methods based on tactile sensing for object identification \cite{BimboResumo}. Many require the full reconstruction of objects \cite{GPAtlas} and can even try to detect specific characteristics and lumps in order to translate them to images \cite{gelsight2}. However, human beings do not need a full 3D reconstruction of the object and typically just need a few grasps to to recognize it \cite{haptic-glance}. If we know what kind of items we have in a bag, we only need a set of features, as the reconstruction is only needed if the object is fully unknown \cite{haptic-perception}.

To obtain the features that characterize an object, one can use the bag-of-features approach \cite{bag-of-features}. In this approach, a two-fingered gripper is used to grasp a fixed object and examine it in height, where each finger of the griper has a sensor pad with 84 sensor cells arranged in 6 columns and 14 rows. The problem with this approach is that it requires it to fully analyse the object, in order to include all the defining features, which might be computationally expensive and slow. Using similar sensors to the ones of the gripper, one can also apply neural networks to the shapes that occur in the pad of sensors. This analysis may allow not only to identify objects but also specific shapes \cite{JimenezNeural}.

The Barrett-Hand has three fingers and two rectangular sensor pads on each one, with a similar distribution to the ones belonging to the previously mentioned gripper. 
The \textit{GraspIt!} Simulator \cite{graspit} can be used to obtain stable grasps for objects. It has been used to compute stable grasps and use them with a real Barrett hand \cite{BlindGrasping},\cite{GraspAdjustment}. The simulator not only provides the grasp qualities but can also provide contact poses where the orientations of the objects, hands and contacts is determined by a quaternion.


Benchmarking in robotic manipulation can be done with the YCB database \cite{ycb-1}. This set consists of different everyday objects, with shape and texture models created from vision algorithms and adapted in order to be used with simulators. These models are compatible with the mentioned \textit{GraspIt!} simulator.



One of the approaches that inspired this project is \cite{feeling} that uses the motor values for each finger and for the hand orientation, of a robotic Barrett Hand. The algorithm then divides each motor's values into 100 bins and applies them onto a novel Bayesian inference update, similar to the Bayes Filter Algorithm in \cite{probabilistic}. This was a simple and effective approach that did not require the values of the tactile sensors, but only of the motors that control both the positions of the fingers and the orientation of the hand. Experiments are made with a fixed object. The wrist is also fixed above the object but the hand can rotate around a vertical axis using one of the motors of the Barrett Hand. This method used 30 orientations of the hand to grasp each object thereby creating tables with the values for each motor. In spite of this, they used the values of the hand orientation motor to determine the correct object. This means that this object has to be fully static or the results will not be valid.
One important aspect to note from that particular work was the implementation of active learning towards the object's exploration. This feature was not applied in any of the works previously referred in this section. This implementation allowed the hand to look for the orientation that had the highest \textit{interestingness} value and therefore let it easily find the grasping location with the most information gain. The main limitation of all the state of the art methods here analysed for robotic grasp object identification is that they are not pose-invariant and, therefore, require the object to be fully static.

There are also several vision algorithms that focus on object identification. A particularly interesting approach uses Point Clouds acquired from a \textit{RGBD} camera and a Point Pair Feature method \cite{pointPair}. This method creates hash tables for distinct objects where the keys are calculated by using the positions and normal angles of two oriented points. This approach can be adapted to robotic grasping, as the oriented points can be defined by the contact locations between the object and the robotic hand.

\section{Approach}

\subsection{Methods}

The core of our approach is based on the use of the position and direction of a contact. This information can be given by 3D force sensors in the robot's fingertip like \cite{optoforce} or \cite{vizzySensors} and it allows us to gather rich information on the identity of the object without knowing the relative pose between object and hand. The method we use is a computer vision algorithm that calculates Point Pair Features \cite{pointPair}, as was mentioned in the previous section. This feature describes relative position and orientation of two oriented points, by using voting schemes. For two contact points $p_1$ and $p_2$ for example, we have positions $m_1$ and $m_2$ and contact forces $n_1$ and $n_2$, respectively. Considering that the distance between two points is defined by $d=m_2-m_1$, the Point Pair feature F is defined as

\begin{equation}
     F(p_1,p_2) = (||d||, \angle(n_1,d),\angle(n_2,d),\angle(n_1,n_2)),
     \label{eq:PPF}
\end{equation}
 
Each PPF (F) is used as a key to an hash table that contains the pairs of points of a model. Note that there will be a hash table for each object and that, in our case, the oriented points will be defined by the contact locations of the robotic hand with the object. This approach is pose invariant since it is based on distances and angles that are independent of the pose of the object. As this method depends on both points and normals, we will call this the \textbf{PN} method.
 
As an alternative, to allow pose invariant haptic object identification in robots that cannot calculate the contact normals, we calculate new hash tables with new keys that use only the distance between points but not the orienting vectors. As this method only requires the points of contact, it will be henceforth be referred to as the \textbf{P} method.

\subsection{Data Collection and processing} 
 
Each grasp can be made in an arbitrary position of the object. There will be as many contact locations as there are fingers of the robotic hand. In the case of the Barrett Hand, there will be three contact locations, originating a vector with values $z=[p_1,p_2,p_3$]=[$(m_1,n_1),(m_2,n_2),(m_3,n_3)]$. It is relevant to consider several data samples for the same grasp as there is some noise associated to the sensor, or even some external disturbances. Because of this, for a single grasp, we extract $N=50$ samples. Each vector of values is used to calculate three keys (pairwise combinations from the set of the 3 contacts) that define features of the object. New features are added to the table and existing keys are updated so that each value of the table corresponds to the number of times that key was produced for that object.

As for the testing part, we need to obtain the vector value to calculate the keys. Each grasp should again produce $N=50$ samples. The set of values for each grasp is now defined by $Z=\{z_1,...,z_{50}\}$. We then check if those keys match the object's hash table and calculate the corresponding number of votes. Each occurrence of a key counts as a vote for that object. The probability that one set of values $Z$ in instance $t$  belongs to object $o_n$ is calculated as

\begin{equation}
    P(Z_t|o_n) = \frac{votes\ for\ object \ o_n}{\sum_i votes\ for\ object\ o_i }.
    \label{eq:vote-percentage}
\end{equation}

As for the vote accumulation, we use a sequential Bayesian Update rule: \hfill
\begin{equation}\label{eq:bayesUpdateF}
P(o_n|Z_t) = \frac{P(Z_t|o_n)P(o_n|Z_{t-1})}{P(Z_t|Z_{t-1})},
\end{equation}
where the denominator is simply a normalization factor so that the sum of the probabilities of the objects adds up to one. The required initial probability is defined for instant $t=0$ as  $P(o_n|Z_0) = P(o_n)= \frac{1}{N}$.

\subsection{Exploration Techniques}

The algorithms described calculate the probability of certainty for a given object. We can now define thresholds that represent how certain the algorithm must be of a classification. The thresholds that will be analysed throughout this project will be $\beta_{thresholds} = \{0.5, 0.6, 0.7, 0.8, 0.9, 0.95, 0.99\}$ and the stop condition is defined by
\begin{equation}
    \begin{array}{l}
        if \  any \  P(o_n|Z_t) > \beta_{threshold} \  then \\
        o' = \argmax\limits_{o_n} P(o_n|Z_t)
    \end{array}
\end{equation}

With no information of the object to be analysed, the first grasp is always in a random position, within the ones trained. We consider that the hand can be placed in arbitrary poses around the object, in a discrete set of L possibilities and each time it grasps the object and calculates the corresponding keys. This means that, when testing, the hand's position can be any of the $L$ possibilities for which the algorithm is trained for. The grasps can either be random in nature, defined by passive exploration, or they can be chosen using active learning. In case of passive exploration, the next grasp is chosen by:
\begin{equation}
    \phi' = Random(0,L),
\end{equation}
where $\phi'$ is one of the possible L hand poses, within the trained orientations, in which the hand will grasp the object. Note that, when using passive exploration, the object can rotate and translate around itself as the hand will always explore randomly and its orientation has no relevance for calculating the keys. This is what makes this approach pose-invariant. When using active exploration, this does not apply anymore as we need to search for the next grasp that provides the most information about the system. The proposed active exploration criteria for this alternative is defined by
\begin{equation}
    \phi' = \argmax\limits_{\phi \in [0,L[} \ P(Z(\phi)|o') - P(Z(\phi)|o'')
\end{equation}
where 
\begin{equation}
    \centering
    \begin{cases}
        o' = \argmax\limits_{o_n \in \{O\}} P(o_n|Z_t)  \\
        o''  = \argmax\limits_{o_n \in \{O\niton o'\}} P(o_n|Z_t)   \\
    \end{cases}
\end{equation}

In the previous equations, \textit{{O}} represents the set of objects, $o'$ is the currently most voted for object and $o''$ is the second most voted object. This means, that the next best grasp is defined by the pose that allows to maximize the difference between the two most likely objects.

\section{Experiments}

\subsection{Robotic Simulator and Data collection}

\begin{figure}[htbp]
\centering{
    \subfloat[Tuna Can]{\includegraphics[width=0.115\textwidth]{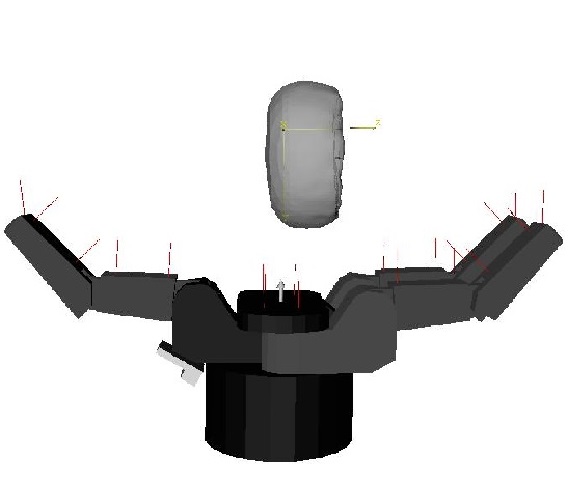}\label{fig:1a}}
    \subfloat[Mug]{\includegraphics[width=0.11\textwidth]{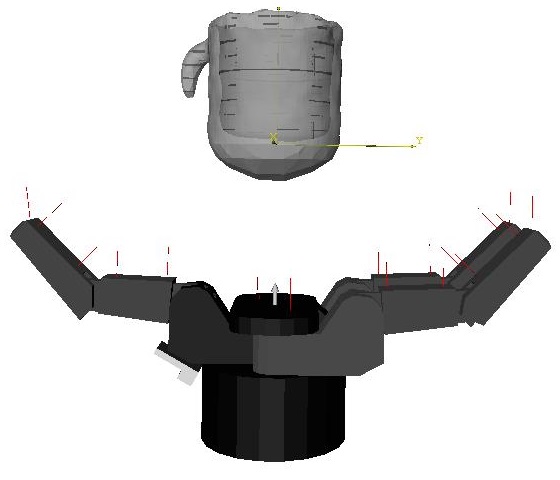}\label{fig:1b}}
    \subfloat[Bowl]{\includegraphics[width=0.11\textwidth]{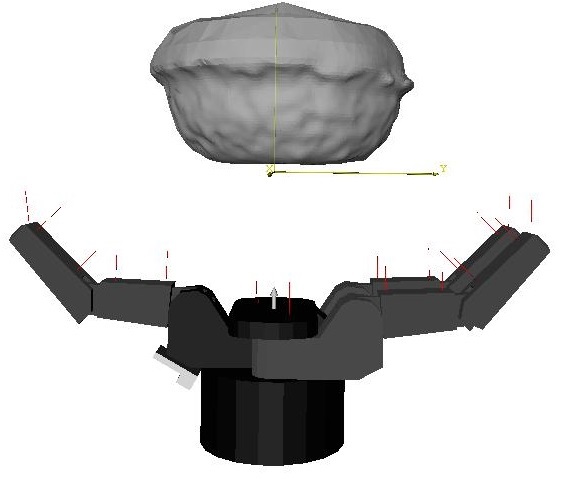}\label{fig:1c}}}
    \\
    \centering{
    \subfloat[Baseball]{\includegraphics[width=0.11\textwidth]{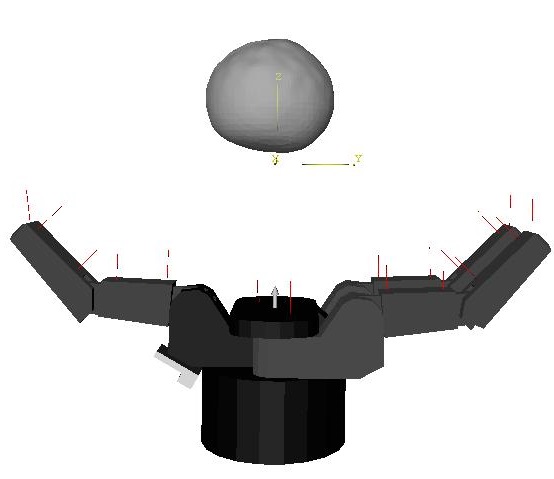}\label{fig:1d}}
    \subfloat[Foam Brick]{\includegraphics[width=0.11\textwidth]{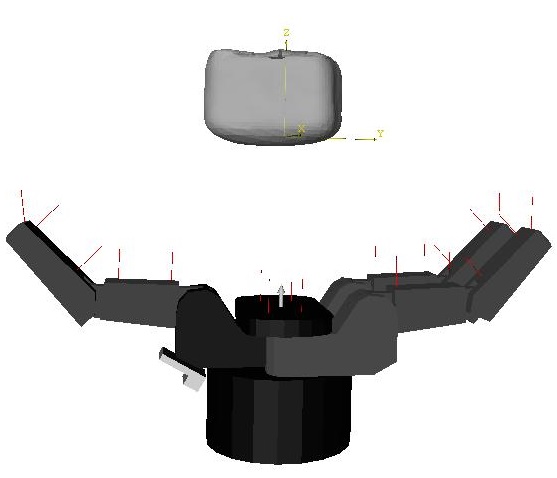}\label{fig:1e}}}
    \caption{Objects used for classification} \label{fig:objects}
\end{figure}

The simulator used in this project was the \textit{GraspIt!} Simulator. The simulator allows the extraction of the values that define the poses (positions and orientations) of the contact locations. With a starting position for each object like is shown in Fig. \ref{fig:objects}, the hand rotates around the vertical (\textit{z}) axis for L=360 degrees, with a one degree step. The contact poses for each object are saved to a distinct file to be processed.


Regarding the objects used for classification, they were chosen from the the YCB object database and uploaded in their \textit{.ply} format. Five objects were chosen and they were labeled as \textit{Tuna Can}, \textit{Mug},\textit{ Bowl}, \textit{Baseball} and \textit{Foam Brick}. All objects are represented in Fig. \ref{fig:objects} with the Barrett Hand, for scale.

\subsection{Creating Hash Tables}
As was mentioned in the previous section, each grasp should produce $N=50$ samples. In a simulator, there is no sensor noise or external disturbances so we must take the distance value and the calculated angles and add random noise $50$ times, as to create $50$ different samples. These samples create the keys for the hash tables that define both methods.

\subsection{Processing grasp values for identification}

After the hash tables are created, we need to process grasp values for object identification. The values associated with each grasp are not retaken as needed for testing, but simply analysed from the previously created file. For each distance value and normal vector, a random noise is added in order to simulate $50$ samples and calculate the corresponding keys, similarly to what we did when creating the table. Note that the noise samples for these keys are different to the ones used when creating the table, as it allows a better approximation to a real life scenario. We then check if those keys exist for each object's hash table and calculate the corresponding object probability, defined in Eq.(\ref{eq:bayesUpdateF}).

\section{Results}

In this section, we present the results for both the \textbf{PN} method and the \textbf{P} method, using both passive and active exploration. Each object was tested $100$ times for each of the methods.

\subsection{Passive learning approach}

\begin{figure}[hbt!]
  \centering
  \includegraphics[width=0.45\textwidth]{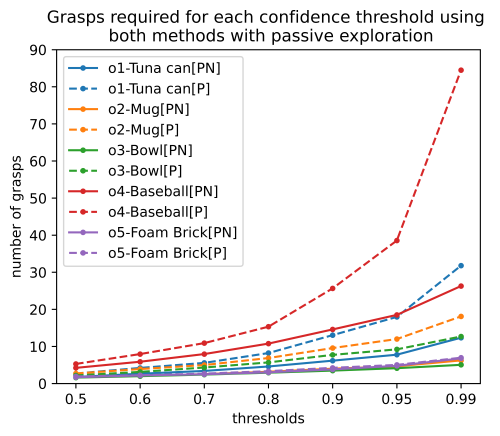}
  \caption{Representation of the average number of grasps for the classification of the objects, for both methods - \textbf{PN} (continuous) and \textbf{P} (dashed)}
  \label{fig:grasps-passive}
\end{figure}

To compare the results between the \textbf{PN} and \textbf{P} methods we show, in Fig. \ref{fig:grasps-passive}, the distribution of the average number of grasps over the $100$ tests for each object, for each confidence threshold. 

It is easy to observe that for the \textbf{PN} method requires, in general, fewer grasps. The exception to the rule is the \textit{Foam Brick} ($o_5$), that presents similar results for both methods. Table \ref{tab:table-passive} shows a more detailed statistical analysis of the results, for the $0.99$ confidence threshold.

\begin{table}[hbt!]
    \centering\caption{Minimum, maximum, average and median values of grasps for a 99\% confidence threshold, for all objects}
    \begin{tabular}{c|cccc|cccc|}
    \centering
    \multirow{2}{*}{} &
          \multicolumn{4}{c|}{\textbf{\textbf{PN}+Passive}} &
          \multicolumn{4}{c|}{\textbf{\textbf{P}+Passive}} \\
        & min & max & avg & med & min & max & avg & med  \\
        \hline
     $o_1$ & 2 & 38 & 12.34 & 11 & 8 & 129 & 31.78 & 26   \\
     $o_2$ & 3 & 16 & 6.23 & 5   & 7 & 47 & 18.09 & 14  \\
     $o_3$ & 1 & 12 & 5.04 & 5   & 1 & 45 & 12.68 & 11.5   \\
     $o_4$ & 2 & 67 & 26.28 & 25 & 13 & 327 & 84.49 & 59  \\
     $o_5$ & 3 & 14 & 6.78 & 7  & 5 & 10 & 6.96 & 7  \\
    \end{tabular}
    \label{tab:table-passive}
\end{table}


\begin{figure}[hbt!]
  \centering
  \subfloat[\textbf{PN} method]{\includegraphics[width=0.24\textwidth]{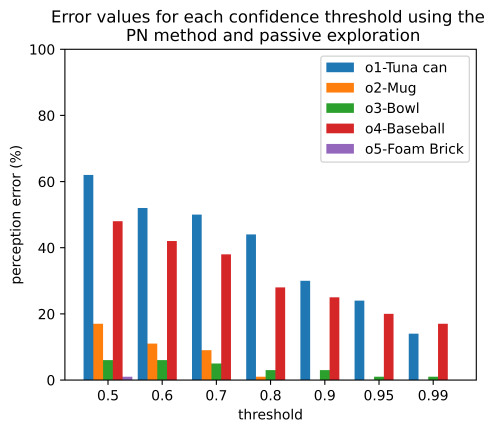}\label{fig:errors-passive-ppf}}
  \hfill
  \subfloat[\textbf{P} Method]{\includegraphics[width=0.24\textwidth]{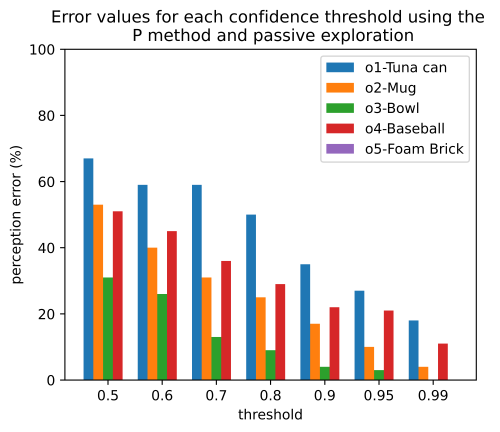}\label{fig:errors-passive-alt}}
  \caption{Percentual perception errors obtained for each object's classification, using Passive learning}
  \label{fig:errors-passive}
\end{figure}

Table \ref{tab:table-passive} shows that both \textbf{P} and \textbf{PN}  methods allowed a single grasp identification for the \textit{Bowl} ($o_3$), which is natural, considering that this is the widest object from the set, thereby creating more distinctive features. For all other objects, the \textbf{PN} method allows the identification in $2$ or $3$ grasps, which is a quite positive result. Instead, the \textbf{P} method requires a minimum of $5$ grasps or more for the identification of other objects.

At the highest confidence threshold, the \textit{Foam Brick} ($o_5$) shows a low grasp number variation and a median grasp value of $7$ for both methods. The minimum amount of grasps, however is $5$ for this threshold. 

The most time consuming object seems to be the \textit{Baseball} ($o_4$), with a maximum value of $327$ grasps for its identification for the \textbf{P} method. It is also the hardest object to identify for the \textbf{PN} method as the maximum number of grasps was $67$. 

Fig. \ref{fig:errors-passive} shows the accuracy of the methods as a function of the confidence threshold. We can observe that the Foam Brick classification error is always close to zero. The characteristics of its features, with large variation in the distances between contact points and flat surfaces, are quite distinct in this dataset. The perception error values for the \textit{Tuna Can} ($o_1$) and \textit{Baseball} ($o_4$) are overall the highest for the \textbf{PN} method. This coincides with the errors for the \textbf{P} method, except for the $0.5$ threshold where the \textit{Mug} slightly surpasses the \textit{Baseball}. For the final threshold, the main difference between the methods is that the error for the \textit{Baseball} lowered more significantly from the previous threshold, for the \textbf{P} method. Even so, this method still presents a perception error of $4\%$ for the Mug, as the \textbf{PN} method was able to totally eliminate any ambiguity for that object.


Both methods proved to be good solutions for the grasp classification task, even for this set of objects that had somewhat similar sizes. Although both seem to have similar error values overall, the \textbf{PN} shows an advantage regarding the number of grasps required for classification.

\subsection{Active learning approach}

\begin{figure}[hbt!]
  \centering
  \includegraphics[width=0.4\textwidth]{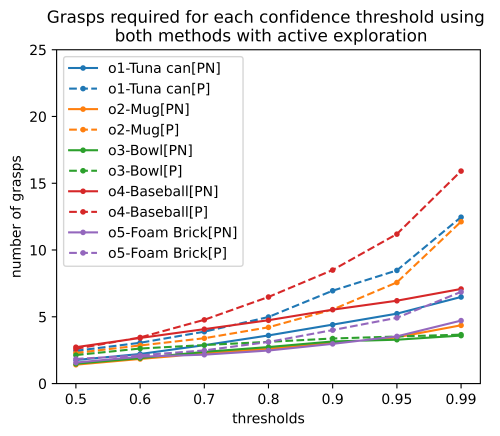}
  \caption{Representation of the average number of grasps for the classification of the objects for both methods - \textbf{PN} (continuous) and \textbf{P} (dashed)}
  \label{fig:grasps-active}
\end{figure}

Fig. \ref{fig:grasps-active} compares the performance of the \textbf{P} and \textbf{PN} methods using active learning. In this case, it is assumed that the relative pose between the hand and the object is known. In our experimental setup, where the object is static, the orientation of the hand is the relevant variable for the exploration. Note that this plot has a much lower scale on the \textit{y-axis}, as this approach required much fewer grasps.


Overall, it is easy to observe that the \textbf{PN} method seems to require less grasps than the \textbf{P} method to identify each of the objects, except for the \textit{Bowl} which presents similar results. The range of possible values also seems a lot higher for the \textbf{P} method as can be verified through Table \ref{tab:table-active}.



\begin{table}[hbt!]
    \centering\caption{Minimum, maximum, average and median values of grasps for all objects, for $\beta_{threshold}$ = 0.99}
    \begin{tabular}{c|cccc|cccc|}
    \centering
    \multirow{2}{*}{} &
          \multicolumn{4}{c|}{\textbf{\textbf{PN}+Active}} &
          \multicolumn{4}{c|}{\textbf{\textbf{P}+Active}} \\
        & min & max & avg & med & min & max & avg & med  \\
        \hline
     $o_1$ & 1 & 16 & 6.48 & 6 & 6 & 38 & 12.46 & 11   \\
     $o_2$ & 3 & 10 & 4.36 & 4 & 5 & 32 & 12.12 & 10  \\
     $o_3$ & 1 & 14 & 3,59 & 3 & 2 & 14 & 3.66 & 3   \\
     $o_4$ & 2 & 19 & 7.08 & 6 & 6 & 73 & 15.91 & 13  \\
     $o_5$ & 2 & 14 & 4.71 & 5 & 5 & 9 & 6.86 & 7  \\
    \end{tabular}
    \label{tab:table-active}
\end{table}

A quick look at the table already allows us to conclude that the active learning greatly lowers the required number of grasps for identification.

\begin{figure}[hbt!]
  \centering
  \subfloat[\textbf{PN} method]{\includegraphics[width=0.24\textwidth]{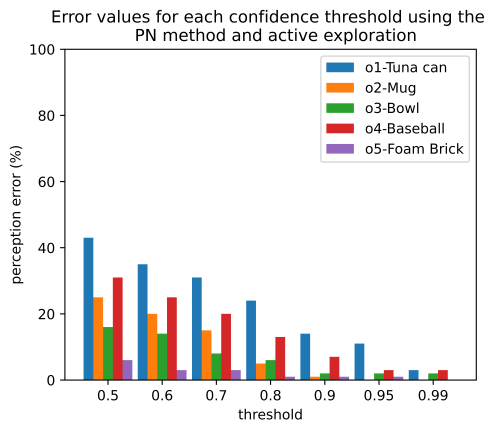}\label{fig:errors-active-ppf}}
  \hfill
  \subfloat[\textbf{P} Method]{\includegraphics[width=0.24\textwidth]{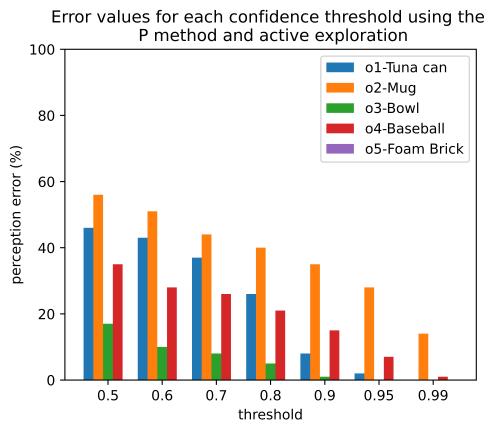}\label{fig:errors-active-alt}}
  \centering\caption{Percentual perception errors obtained for each object's classification, using Active learning}
  \label{fig:errors-active}
\end{figure}


In Fig. \ref{fig:errors-active} we can observe the methods' accuracy for all confidence thresholds. The \textbf{P} method presents an error of $0$ or $1\%$ for most objects at a $0.99$ confidence threshold. However, it has some difficulty identifying the \textit{Mug}, with a classification error of about $15\%$. The error values for the \textit{Mug} are significantly lower for the \textbf{PN} method, having reached $0\%$ for the $0.95$ threshold, which is a much better result than for the \textbf{P} approach. This seems to be the hardest object for the \textbf{P} method with classification errors always growing with the decrease of the confidence threshold.

As in the passive exploration case, for the \textbf{P} method, the \textit{Foam Brick} has a classification error of 0\%, for all analysed confidence thresholds. The same occurrence does not happen for the \textbf{PN} method, as the perception error of the \textit{Foam Brick}  only reaches $0\%$ in the $0.99$ confidence threshold. 

\subsection{Comparing Passive and Active Approaches}

Regarding the error values for the \textit{Tuna Can} and the \textit{Baseball}, we can compare Fig. \ref{fig:errors-passive}(a) and Fig. \ref{fig:errors-active}(a) to see that they are much lower with active learning for all thresholds, having decreased $\approx 20\%$ for the $0.5$ certainty level. 
When comparing Fig. \ref{fig:errors-passive}(b) and Fig. \ref{fig:errors-active}(b), related to the \textbf{P} method, we can see that the perception errors for the \textit{Tuna Can} and the \textit{Baseball} are lower overall, and especially so in the $0.99$ threshold. However, the \textit{Mug} presents a much higher error for the passive exploration alternative, higher even than for those two objects. This can happen due to the fact that there are some features common to many objects but more abundant in one. If the grasps continue to obtain those features, they are likely to misclassify the object. For similar reasons, if we are using active learning but the most likely object at the moment is not the correct one, it can provoke a series of grasps that calculate features more abundant for another object. Since the active learning algorithm ends up choosing a similar series of grasps for the same most likely objects, the classification follows the same path several times. Random exploration allows avoiding that path, which apparently worked well to identify the \textit{Mug}.


\subsection{Overall results}

\begin{figure}[!h]
  \centering
  \includegraphics[width=0.45\textwidth]{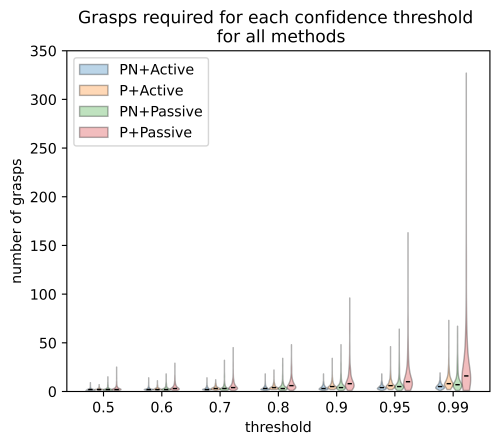}
  \caption{Violin plot showing the grasp distribution and the median required grasps to identify each object, for all methods}
  \label{fig:grasps-all}
\end{figure}

In Fig. \ref{fig:grasps-all} we present a summary of the overall results in the form of a violin plot. This plot allows us to evaluate the overall distribution of the number of grasps for all methods, as a function of the confidence threshold. As expected from the previous analysis, the active \textbf{PN} method requires less grasps overall for object identification than the \textbf{P} Method, with median values of $5$ and $8$ grasps, respectively, as can be verified through Table \ref{tab:table-total}.

\begin{figure}[hbt!]
  \centering
  \includegraphics[width=0.45\textwidth]{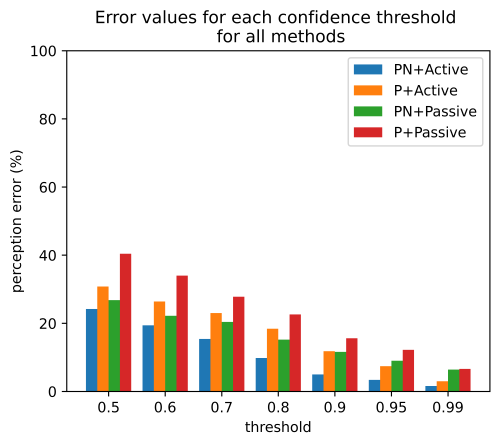}
  \caption{Percentual perception error for all methods}
  \label{fig:errors-all}
\end{figure}

Interestingly, even without active exploration, the \textbf{PN} method requires less grasps to classify the objects than the \textbf{P} method. The combination of the \textbf{P} method with the passive exploration was by far the one that required the most grasps for identification. On top of this, it also proved to be the one with the highest perception errors, as seen in Fig. \ref{fig:errors-all}.

\begin{table}[hbt!]
    \centering\caption{Minimum, maximum, average and median values of grasps for all methods, for $\beta$  = 0.99}
    \begin{tabular}{c|cccc|}
    \centering
      {}   & min & max & avg & med  \\
      \hline
     \textbf{PN}+Active & 1 & 19 & 5.24 & 5 \\
     \textbf{P}-Active & 2 & 73 & 10.20 & 8    \\
     \textbf{PN}-Passive & 1 & 67 & 11.33 & 7    \\
     \textbf{P}-Passive & 1 & 327 & 30.80 & 16   \\
    \end{tabular}
    \label{tab:table-total}
\end{table}

Comparing the perception errors, we see that the passive \textbf{PN} method  still produced lower errors that the active \textbf{P} method, up until the $0.9$ confidence threshold. Even though the error for the passive \textbf{PN} method lowered until the $0.99$ threshold, it lowered less than the active \textbf{P} method. If we analyse only passive exploration, we can conclude that the \textbf{PN} method generates lower error values overall, although both methods get very similar results for the last certainty threshold.

For all approaches and all objects, we can see that the execution time gradually increased as the perception error decreased. Results showed that actively exploring an object leads to faster classifications and lower perception errors. Also, the range of grasps required for classification is greatly improved for most objects.

\section{Conclusions}

In this paper we have presented two methods to classify objects from sequences of grasps without requiring knowledge about the relative pose between the hand and the object along the sequence. This permits that the object moves between grasps, which is an important issue in a practical point of view. The first method (\textbf{PN}) uses the information about the position and orientation of the contacts between the fingers and the object, relative to the hand reference frame. The second method (\textbf{P}) uses only the position, accounting for robots with less haptic sensing abilities. Experiments have shown that both methods demonstrate a good ability to recognise objects in a limited set with good accuracy and using a small number of grasps. The \textbf{PN} method, because it uses richer features, naturally achieves better performance, reducing the number of needed grasps to about half of the P method, on average. Considering the case where the relative pose between hand and object can be measured (e.g. with external sensors) we propose a method that uses active exploration to further reduce the number of grasps. Here, the \textbf{PN} method still shows better results, but the difference to the \textbf{P} method is smaller. This suggests that, while contact orientation sensing makes a difference in the pose-free case, simpler contact sensing can be sufficient to the case where we can measure the pose of the object relative to the hand. Future work will focus on a more thorough characterisation of our methods, using more objects, more degrees of freedom in the hand movement, and more natural environments, both in simulation and with a real robot hand. Nevertheless, this paper presents for the first time a pose-free method for haptic recognition that will improve the efficiency of haptic sensing methods for robots in the recognition of objects in natural conditions.

\printbibliography

\end{document}